\documentclass{sig-alternate-05-2015}

\usepackage{graphicx}
\usepackage{amssymb}
\usepackage{times}
\usepackage{amsmath}
\usepackage{booktabs}
\usepackage{mathrsfs}
\usepackage{epstopdf}

\usepackage{times}
\usepackage{graphics}
\usepackage{epstopdf}
\usepackage{epsfig}
\usepackage{subfig}
\usepackage{algorithm}
\usepackage{algorithmic}
\usepackage{tabularx}
\usepackage{caption}
\usepackage{algorithm}
\usepackage{extarrows}
\usepackage{algorithmic}
\newcommand{\eat}[1]{}
\usepackage{verbatim}
\begin{document}

\setcopyright{acmcopyright}

\conferenceinfo{WOODSTOCK}{'97 El Paso, Texas USA}

\title{Zero-Shot Hashing via Transferring Supervised Knowledge}

\author{
Yang Yang$^1$,
Weilun Chen$^1$,
Yadan Luo$^1$,
Fumin  Shen$^1$,
Jie Shao$^1$ and
Heng Tao Shen$^{2,1}$\\
\affaddr{$^1$University of Electronic Science and Technology}\\
\affaddr{$^2$The University of Queensland}\\
\email{\{dlyyang,chenweilunster\}@gmail.com,~lyd\_uestc@163.com}\\
\email{fumin.shen@gmail.com,~shaojie@uestc.edu.cn,~shenht@itee.uq.edu.au}
}

\maketitle
\begin{abstract}
Hashing has shown its efficiency and effectiveness in facilitating large-scale multimedia applications. Supervised knowledge (\emph{e.g.}, semantic labels or pair-wise relationship) associated to data is capable of significantly improving the quality of hash codes and hash functions. However, confronted with the rapid growth of newly-emerging concepts and multimedia data on the Web, existing supervised hashing approaches may easily suffer from the scarcity and validity of supervised information due to the expensive cost of manual labelling. In this paper, we propose a novel hashing scheme, termed \emph{zero-shot hashing} (ZSH), which compresses images of ``unseen'' categories to binary codes with hash functions learned from limited training data of ``seen'' categories. Specifically, we project independent data labels (\emph{i.e.}, 0/1-form label vectors) into semantic embedding space, where semantic relationships among all the labels can be precisely characterized and thus seen supervised knowledge can be transferred to unseen classes. Moreover, in order to cope with the semantic shift problem, we rotate the embedded space to more suitably align the embedded semantics with the low-level visual feature space, thereby alleviating the influence of semantic gap. In the meantime, to exert positive effects on learning high-quality hash functions, we further propose to preserve local structural property and discrete nature in binary codes. Besides, we develop an efficient alternating algorithm to solve the ZSH model. Extensive experiments conducted on various real-life datasets show the superior zero-shot image retrieval performance of ZSH as compared to several state-of-the-art hashing methods.
\end{abstract}


\section{Introduction}
Hashing~\cite{wang2014hashing} is a powerful indexing technique for enabling efficient retrieval on large-scale multimedia data, such as image~\cite{shen2015supervised,Shen_2015_ICCV,yang2015visual} and video~\cite{cao2012submodular}. Specifically, in order to achieve shorter response time and less computational cost, hashing encodes high-dimensional data into compact binary codes (\textit{i.e.,} 0 or 1) substantially. In this way, data can be compactly stored and Hamming distances can be efficiently calculated with bit-wise XOR operations. Because of its impressive capacity of dealing with ``curse of dimensionality'' problem, hashing has been extensively employed in various real-world applications, ranging from multimedia indexing~\cite{hu2014iterative,zhu2013linear,yang2014exploiting,yang2015robust} to multimedia event detection~\cite{petrovic2010streaming}.

\begin{figure}[t]
  \centering
  \includegraphics[width=\linewidth]{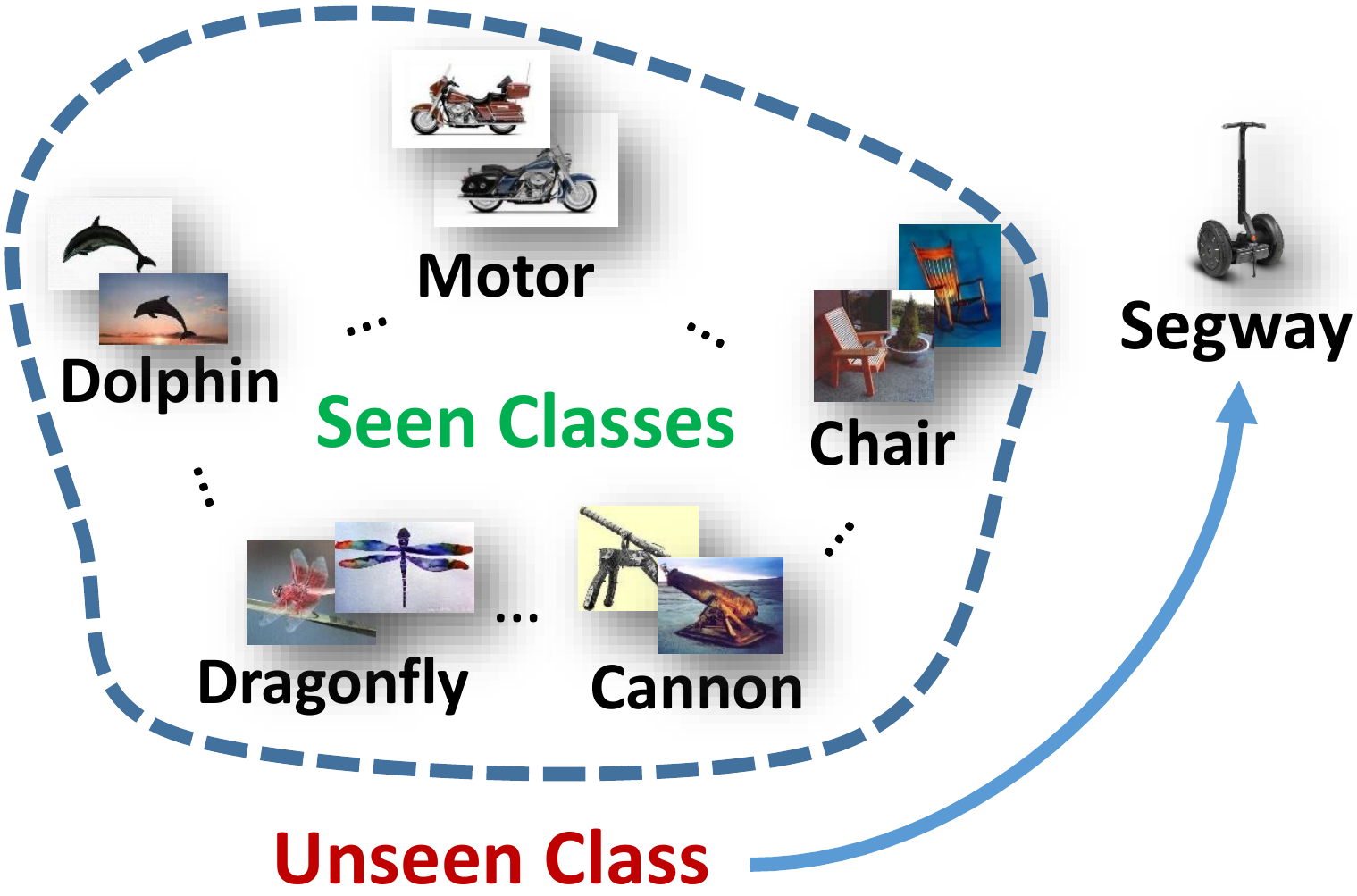}\\
  \caption{An illustration of newly-emerging concepts and images unseen to the existing learning systems.}
  \label{fig:unseen}
\end{figure}

There are mainly two branches of hashing, \textit{i.e.,} data-independent hashing and data-dependent hashing. For data-independent hashing, such as Locality Sensitive Hashing~\cite{gionis1999similarity}, no prior knowledge (\emph{e.g.}, supervised information) about data is available, and hash functions are randomly generated. Nonetheless, huge storage and computational overhead might be cost since more than $1,000$ bits are usually required to achieve acceptable performance. To address this problem, research directions turn to data-dependent hashing, which leverages information inside data itself. Roughly, data-dependant hashing can be divided into two categories: unsupervised hashing (\emph{e.g.}, Iterative Quantization~\cite{gong2011iterative} and Sparse Mutli-Modal Hashing~\cite{wu2014sparse}), and (semi-)supervised hashing (\emph{e.g.}, Supervised Hashing with Kernels~\cite{liu2012supervised}, Supervised Discrete Hashing~\cite{shen2015supervised}, Discrete Graph Hahsing~\cite{liu2014discrete} and Semi-Supervised Hashing~\cite{wang2010semi}). In general, supervised hashing usually achieves better performance than unsupervised ones because supervised information (\emph{e.g.}, semantic labels and/or pair-wise data relationship) can help to better explore intrinsic data property, thereby generating superior hash codes and hash functions.

Along with the explosive growth of Web data, traditional supervised hashing methods have been facing an enormous challenge, \emph{i.e.}, the generation of reliable supervised knowledge cannot catch up with the rapid increasing speed of newly-emerging semantic concepts and multimedia data. In other words, due to the expensive cost of manual labelling (time-consuming and labor-intensive), sufficient labelled training data is usually not timely available for learning new hash functions that can accurately encodes data of new concepts. As illustrated in Figure~\ref{fig:unseen}, within the ``seen'' zone, where images are attached with known categories, existing supervised hashing algorithms may perform well because they are fed with correct guidance. However, outside the seen area, supervised hashing algorithms may easily fail to generalize to data of new categories that they never observe, \textit{e.g.,} $segway$, a two-wheeled, self-balancing, battery-powered electric vehicle. Moreover, most of current approaches use supervised information in the form of either 0/1 semantic labels or pair-wise data relationship for guiding the learning process, which implies that precious correlation among label semantics are inevitably ignored. One straightforward consequence of the semantic independency is that each category can neither learn from other relevant categories nor distribute its own supervised knowledge to other seen classes and/or even those unseen ones.

The aforementioned disadvantages motivate us to consider whether we can encode images of ``unseen'' categories into binary codes with hash functions learned from limited training samples of``seen'' categories? The key challenge of achieving this goal is how to set up a tunnel to transfer supervised knowledge between ``seen'' and ``unseen'' categories. In recent years, zero-shot learning (ZSL)~\cite{palatucci2009zero,socher2013zero,castanon2015efficient,norouzi2013zero} has been widely recognized as a way to deal with this problem. The ZSL paradigm aims to learn a general mapping from the feature space to a high-level semantic space, which helps avoid rebuilding models for unseen categories with extra manually labelled data.  ZSL is mostly achieved by using class-attribute descriptors to bridge the semantic gaps between low-level features and high-level semantics, where new categories are thus learned using only the relationship between attributes and categories. However, most of existing attribute based ZSL methods still suffer from: (1) erroneous guidance derived from imprecise or incomplete human-labelled attributes~\cite{jayaraman2014zero}, which is usually due to the lack of expertise or mislabeling by annotators, etc.; (2) diminishing of discrimination for pre-defined attributes when confronted with dataset shift~\cite{lampert2014attribute,parikh2011relative}.

Recently, mining other auxiliary datasets has been shown to be helpful to tackle the zero-shot learning problem. For instance, with a huge corpus such as Wikipedia, one can obtain word embeddings that capture distributional similarity in the text corpus~\cite{turian2010word}, such that similar words can be located in similar place. During the learning phase, visual modality can be grounded by the word vectors, and such knowledge can thus be transferred into the learned model. Inspired by this, many approaches choose to utilize auxiliary modalities to help address the zero-shot tasks. Socher et al.~\cite{socher2013zero} uses word embedding as supervision in order to detect novel categories and perform classification accordingly. Frome et al.~\cite{frome2013devise} adopts a similar manner, which connects raw features and word embedding space using the dot-product similarity and hinge rank loss. In the hashing domain, however, the zero-shot problem has rarely been studied.

As previously analyzed, with the newly-emerging concepts and multimedia data, we are in urgent demand of a reliable and flexible hash function that can be adopted to hash images of unseen categories. In this work, we propose a novel hashing scheme, termed \emph{zero-shot approach} (ZSH). Inspired by the superior capacity of the word embedding for capturing the semantic correlations among concepts, we map mutually independent labels into a semantic-rich space, where supervised knowledge of both seen and unseen labels can be completely shared. This strategy helps to encode images of unseen categories without any assistance of visual observation in those unknown classes. Besides, even though we cannot retrieve images of exactly the same category, semantically related objects can be returned. Moreover, we recognize the problem of semantic shift caused by off-the-shelf embedding. The embedded space is then rotated to make the hash functions more generalized to images of unseen categories. To further improve the quality of hash functions, we also preserve local structural property and discrete nature in binary codes. We summarize our main contributions as below:

\begin{itemize}
\item We address the problem of employing training data of seen categories to learn reliable hash functions for transforming images of unseen categories into binary codes. We propose a novel zero-shot hashing, which bridges gaps between originally independent labels through a semantic embedding space. To the best of our knowledge, this is one of the first works that study the problem of hashing data from newly-emerging concepts with limited seen supervised knowledge. Extensive experiments on various multimedia data collections validate the efficacy of our proposed ZSH.
\item We devise an effective strategy for transferring available supervised knowledge from seen classes to unseen classes. In particular, we transform labels into a word embedding space, where semantic correlations among labels can be quantitatively measured and captured. In this way, unseen labels can leverage the well-established mapping from its semantically close seen categories. For instance, \emph{segway} may learn from \emph{bicycle} and \emph{automobile}.
\item Since the initial semantic embedding is from an off-the-shelf word embedding space, which may bring in severe semantic shift between categories and the original visual feature. To alleviate the potential influence, we propose to further rotate the embedding space to better fit the underlying feature characteristics, thereby narrowing down the semantic gap effectively.
\item In order to generate more reliable hash functions, we propose to improve the intermediate binary codes of training data by exploring underlying data properties. Concretely, we impose discrete constraints on binary codes during the code learning process as well as preserve data local structure, \emph{i.e.}, if two datums share similar representations in the original space, they are supposed to be close to each other in the learned Hamming space.
\end{itemize}

The rest of this paper is organized as follows. In Section 2, we briefly review some related work on hashing and zero-shot learning. In Section 3, we will elaborate our approach with details, together with our optimization method and an analysis of the algorithm. With extensive experiments, various results on various different datasets will be reported in Section 4, followed by the conclusion of this work in Section 5.

\begin{figure*}[!ht]
  \centering
  \includegraphics[width=1\linewidth]{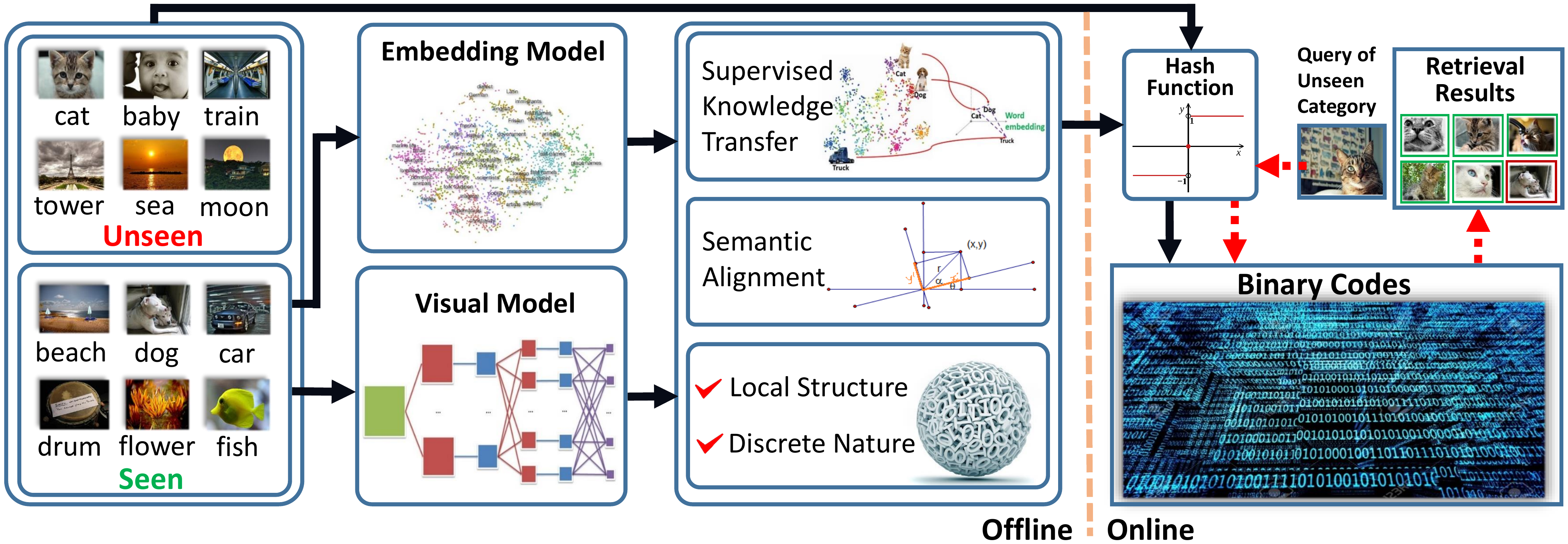}\\
  \caption{An illustration of the overall architecture of the proposed zero-shot hashing framework.}
  \label{fig:framework}
\end{figure*}

\section{Related work}
In this section, we aim to clarify the relationship between our work and other researches, due to the constraints of space, we cannot completely elaborate every detail of previous literature.

\subsection{Zero-Shot Learning}
Learning with no data, \textit{i}.\textit{e}., zero-shot learning has been proved to be an efficient approach to tackle the increasing difficulty posed by insufficient training examples. Many approaches have been proposed to solve this problem by using an intermediate layer to represent an image. Specifically, with visual attributes or other semantic abundant descriptors, a novel image can thus be defined as the relationship between category and intermediate representation. In the work ~\cite{farhadi2009describing} by Farhadi, he leverages attributes as a way to classify unseen objects by describing them with attributes. The work ~\cite{larochelle2008zero} by Larochelle named \textit{Zero-data Learning of New Tasks} has also proven to be useful when predict categories that are not shown in the training dataset. Recently, learning novel images with auxiliary datasets (\textit{e.g.}, leveraging textual relationship in a large corpus) has been shown to be powerful at doing zero-shot tasks. Learning the correlations between concepts, the label of a novel example omitted from training set can be reasonably inferred. Renown works include Socher's work~\cite{socher2013zero} \textit{Zero-shot Learning Through Cross-Modal Transfer}, which uses label embedding to detect unseen classes and makes semantically reasonable deduction. \textit{DEVISE}~\cite{frome2013devise} also uses the same scheme as ~\cite{socher2013zero}, but with a different language modal and a different loss function to connect two modalities. However, all above methods are limited to classification or prediction scenario. To our best knowledge, we are the first one to handle the zero-shot retrieving problem, \textit{i}.\textit{e}., hash novel images that were not observed. By adopting a natural language model~\cite{huang2012improving} pre-trained with a large corpus from Wikipedia, we precisely capture the correlations between different words, and thus hash unseen images into correct Hamming space.

\subsection{Hashing}
This subsection overviews fast search with binary codes using hashing technique. Similarity search is a challenge of pursuing data points of smallest distance in a large scale database. The easiest hashing scheme is dubbed Local Sensitive Hashing~\cite{gionis1999similarity}, which designs hashing function with no prior knowledge of the data distribution. However, such hashing methods require significantly large code length to achieve an acceptable performance, generating large overheads in a database. To address this problem, learning to hash comes as a trend. Unsupervised hashing methods mine the statistic distributional information in the database, generating an optimal hashing function to preserve the similarity in the original space. Classical algorithms such as Spectral Hashing (SH)~\cite{weiss2009spectral}, solves binary codes to preserve the Euclidean distance in the database; Inductive Manifold Hashing (IMH)~\cite{shen2013inductive}, adopts manifold learning techniques to better model the intrinsic structure embedded in the feature space; Iterative Quantization~\cite{gong2011iterative}, focuses on minimizing quantization error during unsupervised training. Considering a real-world database is commonly described by multiple modalities, such as visual features (\emph{e.g.}, Caffe~\cite{jia2014caffe}) or textual information (\emph{e.g.}, image captions, lyrics), Sparse Multi-Modal Hashing~\cite{wu2014sparse} utilizes information of at least two different resources to achieve promising performance. Since the unsupervised way is guided with little human-level knowledge, supervised hashing have been proposed to use supervision information to learn binary codes. Hashing techniques in this category have been emerging continuously in recent years, representative methods include Kernel Supervise Hashing (KSH)~\cite{liu2012supervised}, Minimal Loss Hashing (MLH)~\cite{norouzi2011minimal}, Supervise Discrete Hashing (SDH)~\cite{shen2015supervised}, Latent Factor Hashing (LFH)~\cite{zhang2014supervised} as well as the recently proposed Column Sampling Based Discrete Supervised Hashing (COSDISH) \cite{kang2016column}, \textit{etc.} Using supervision information, these hashing schemes perform better than unsupervised ones. Recently, with the rising of deep learning, image hashing using large convolutional neural network has also be shown to be effective in hashing domain~\cite{xia2014supervised}. By using hidden layers to represent images as feature vectors that are optimal for binary codes generation, hashing performance can be augmented greatly.

Admittedly, hashing algorithms have successfully tackled the ``curse of dimensionality'' in terms of fast search, however, what if we want to achieve data-dependent performance while no training example is provided? All above hashing methods fail to generalize to ``unseen'' categories, limiting in the ``seen'' area where every category correspond to at least one training image. Besides, as the database changes everyday, re-training hashing function frequently can be expensive, further prevents their practical usage in large dynamic real-world databases. Based on tabove analysis, a hashing method that can perform well on unseen data draws a strong need, thus the orientation of zero-shot hashing is quite obvious.

\section{Zero-Shot Hashing}
In this section, we elaborate our proposed zero-shot hashing (ZSH). We firstly present a formal definition of hashing in zero-shot scenario, and then depict the details of ZSH, including a brief introduction of overall framework, supervision transfer, semantic alignment as well as hashing model. Finally, we introduce the optimization process and algorithm analysis.

\subsection{Problem Definition}
Suppose we are given $n$ training images ${X} = \left[ {x_1,x_2, \ldots ,x_n} \right] \in {\mathbb{R}^{d \times n}}$ labeled with a seen visual concept set $\mathcal{C}$, where $d$ is the dimensionality of visual feature space. Denote $Y = \left[ {{y_1},{y_2}, \ldots ,{y_n}} \right] \in {\{ 0,1\} ^{c \times n}}$ is the binary label matrix, where $y_i\in\{0,1\}^{c\times 1}$ is the label vector of the $i$-th sample $x_i$ and $c$ is the number of seen classes in $\mathcal{C}$. Different from conventional supervised hashing scenario, where both testing data and training data are associated with the same concept set, \textit{i.e.}, $\mathcal{C}$, we intend to cope with the situation where testing data and training data share no common concepts. In other words, testing data (denoted as $X^{(u)}$) belongs to an ``unseen'' category set $\mathcal{C}^{(u)}$, \textit{i.e.}, $\mathcal{C}^{(u)}\cap\mathcal{C}=\emptyset$. Using only the training images $X$ where no training samples of the ``unseen'' categories in $\mathcal{C}^{(u)}$ are available, our goal is to learn a hash function $f:\mathbb{R}^d\mapsto \{-1,1\}^{l\times 1}$, which can map images belonging to both $\mathcal{C}^{(u)}$ and $\mathcal{C}$ from original visual feature space to $l$-bit binary codes. The learned hash function $f$ not only guarantees that the binary codes of semantically relevant objects have short Hamming distances, but also generalizes well to the testing data belonging to the unseen categories, even though no training data are utilized in the training phase.

\subsection{Overall Framework}

The flowchart of our overall framework is illustrated in Figure~\ref{fig:framework}. As we can see, there are two stages: the offline phase and the online phase. In the offline phase, suppose only images of a limited number of categories are visible to our system. We firstly extract their visual feature features through a convolutional neural network. At the same time, we use a NLP model to transform seen labels into a semantic-rich embedding space, where each label is represented by a real-valued vector. With the embedded semantics, the relationships among both seen and unseen categories can be well captured and characterized. Instead of $0/1$-form label vector, ZSH supervises the learning of hash functions with the embedded semantic vectors to transfer supervised knowledge. We further rotate the off-the-shelf embedding space to better align with the low-level visual feature space. Meanwhile, ZSH preserves local structural information and discrete nature of the intermediate binary codes to improve hash functions. Finally, we use the learned hash functions to transform all the images in the database into binary codes for subsequent retrieval. In the online phase, when a new query image of any unseen category comes, we encode the new image into binary code following the same mapping and retrieve images that are close to this query in the Hamming space.

\subsection{Transferring Supervised Knowledge}
\begin{figure}[t]
  \centering
  \includegraphics[width=1.02\linewidth]{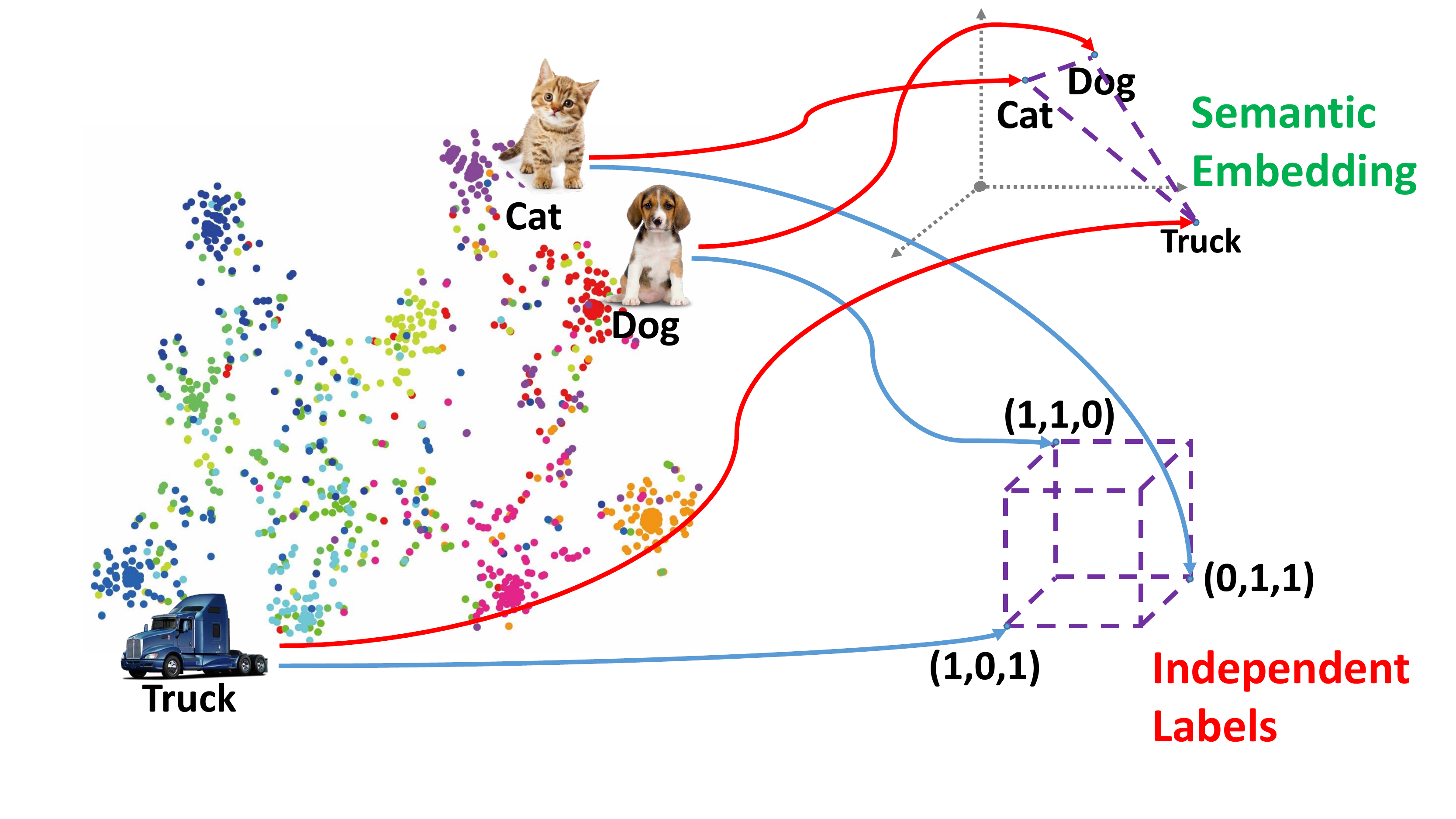}\\
  \caption{An illustration of independent $0/1$-form labels v.s. word embedding.}
\label{fig:emb}
\end{figure}

In general, most of existing supervised hashing algorithms may retrieve relevant results of queries in the seen categories since there are supervised information for understanding the queries. Nevertheless, when the hashing systems have no knowledge of certain unseen classes, query images from these classes will be probably be misunderstood, thereby leading to inaccurate search. One of the main causes is that the supervised information is in the form of $0/1$-form label vectors or pair-wise data relationship, which implicitly makes labels independent to each other and omits the inherent correlation among their high-level semantics (\textit{e.g., }\textit{cat} is as different from \textit{truck} as from \textit{dog}). As illustrated in Figure \ref{fig:emb}, using independent labels, each object will be mapped to an independent vertex of a hypercube, and the distance between any two categories will be the same. In order to address such disadvantage, we propose to connect label semantics by taking advantage of the superior ability endowed by neural language processing techniques. Specifically, as illustrated in Figure \ref{fig:emb}, we map independent labels into a word embedding space, where semantic correlations among labels can be quantitatively measured and captured. Therefore, unseen labels can leverage the well-established mapping from its semantically close seen categories. For example, in the embedding space, \textit{cat} and \textit{dog} will be close to each other, hence even the hashing systems may never observe any \textit{cat} images, they can still gain some useful clues from the supervised knowledge of \textit{dog}. We adopt the language model~\cite{huang2012improving} pre-trained using free Wikipedia text. This model leverages not only local information but also global document context, therefore shows superior performance over other competitive approaches. Every category is embedded into a $50$-d word vector\footnote{In practice, we find that by setting word vector to unit length, retrieval performance can be augmented with no distortion of the cosine similarities, thus we empirically normalize word vector to be unit length.}. In the subsequent part, we consistently denote the embedded label matrix as $Y$ for brevity.

\subsection{Semantic Alignment}
Note that the transformed supervised knowledge from the off-the-shelf embedding space may potentially deviate from the underlying semantics of the image data due to the problems such as domain difference, semantic shift, semasiological variation. This will inevitably jeopardize the whole learning process in our proposed model. In order to prevent this issue, we propose to a semantic alignment strategy, which actively aligns the initial embedding space with the distributional properties of low-level visual feature. In particular, we seek for certain transformation $R\in\mathbb{R}^{c\times c}$ matrix with orthogonal constraint $R^TR=I_c$ to rotate the embedding space to $R^TY$. Recall that we intend to use the amendatory supervised knowledge to guide the learning of high-quality hash codes and hash functions, therefore, we minimize the following error:
\begin{equation}
\left\| {R^TY - {W^T}B} \right\|_{F}^{2},
\end{equation}
which $W^T\in R^{c\times l}$ is the mapping matrix from binary codes $B\in \{-1,1\}^{l\times n}$ to the supervised information. $l$ is the code length. The benefit of the above formulation is that it can help to narrow down the semantic gap between binary codes and the supervised knowledge.

\subsection{Hashing Model}
For convenience, we firstly recap some previous settings here. Suppose we have $n$ training samples $X = \left [  x_{1},x_{2},\cdots ,x_{n} \right ]\in \mathbb{R}^{d\times n}$. For brevity, we denote the corresponding embedded label knowledge as $Y=\left [y_{1},y_{2},\cdots,y_{n} \right ]\in\mathbb{R}^{p\times n}$. Our ultimate target is to learn a set of hash functions from ``seen'' training data $X$ supervised by $Y$, enabling generating high-quality binary codes for data of ``unseen'' categories. Meanwhile, the quality of hash functions may heavily rely on the reliability of the intermediate binary codes of training data. In other words, the model is supposed to simultaneously well control both hash functions and hash codes. To achieve the above goals, we propose the following model:

\begin{equation}
\begin{split}
\min \limits_{f,W,B,R}~&{\left\| {R^TY - {W^T}B} \right\|_{F}^{2}} + \lambda {\left\| W \right\|_{F}^{2}} + \alpha \left\| {f(X) - B} \right\|_F^2
\\&+\beta \left\| {P} \right\|_F^2+ \gamma \sum\limits_{i = 1}^n {\sum\limits_{j = 1}^n {{S_{ij}}\left\| {f({x_i}) - f({x_j})} \right\|_F^2} } \\
&\textrm{s.t.}~B\in \{-1,1\}^{l\times n}\wedge R^TR = I_c,
\end{split}
\label{eq:zsh}
\end{equation}
where $R\in \mathbb{R}^{c\times c}$ is the semantic alignment matrix. $W\in \mathbb{R}^{l\times c}$ is the mapping matrix from binary codes to supervisory information. $B = \left [b_{1},b_{2},\cdots,b_{n} \right ]\in\left \{ -1,1 \right \}^{l\times n}$ denotes the binary codes of $X$, where $b_i$ is the binary codes of the $i$-th sample $x_{i}$. $I_c$ is a diagonal matrix of size $c\times c$. $\|\cdot\|_F$ denotes the Frobenius norm of a matrix. $\lambda>0,\alpha>0,\beta>0$ and $\gamma>0$ are balancing parameters. $f:\mathbb{R}^{m\times 1}\rightarrow\mathbb{R}^{l\times 1}$ define a hash function from a non-linear embedded feature space to the desired Hamming space:
\begin{equation}
f(x)=P^T\phi(x),
\label{eq:f}
\end{equation}
where $f(X)=[f(x_1),f(x_2),\ldots,f(x_n)]$. $P\in \mathbb{R}^{l\times m}$ is the transformation matrix. Following the successful practice for learning hash functions in~\cite{liu2012supervised}, we employ kernel mapping to handle the potential problem of linear inseparability:
\begin{equation}
\phi(x) \!= \!\left[\exp(-\frac{\|x\!-\!a_1\|^2}{\delta}),\ldots,\exp(-\frac{\|x\!-\!a_m\|^2}{\delta})\right]^T.
\end{equation}
where $\{a_i\}|_{i=1}^m$ are $m$ anchors randomly sampled from $X$ and $\delta$ is the bandwidth parameter.

Note that we keep the discrete constraint on the variable $B$ to prevent information loss of binary codes to the greatest extent. The term $\gamma \sum\nolimits_{i = 1}^n {\sum\nolimits_{j = 1}^n {{S_{ij}}\left\| {f({x_i}) - f({x_j})} \right\|_F^2} } $ in Eq.~(\ref{eq:zsh}) preserves local structural information of training data, \textit{i.e.}, if two samples are similar in the original feature space (large $S_{ij}$), then they are enforced to share similar binary codes in the Hamming space.

In the next part, we introduce an efficient alternating algorithm to optimize our zero-shot hashing model.

\subsection{Optimization}
We first rewrite the model in matrix form as follows:
\begin{equation}
\begin{split}
\min \limits_{P,W,B,R}~&{\left\| {R^TY - {W^T}B} \right\|_{F}^{2}} + \lambda {\left\| W \right\|_{F}^{2}} + \alpha \left\| {P^T\phi(X) - B} \right\|_F^2
\\&+\beta \left\| {P} \right\|_F^2+ \gamma Tr(P^T\phi(X)L\phi(X)^TP)
\\ &\textrm{s.t.}~B\in \{-1,1\}^{l\times n} \wedge R^TR = I,
\end{split}
\label{eq:zsh2}
\end{equation}
where $\phi(X)=[\phi(x_1),\phi(x_2),\ldots,\phi(x_n)]$ and the Laplacian matrix $L$ is computed as:
\begin{equation}
L = D-S,
\end{equation}
where $D$ is a diagonal matrix with its $i$-th diagonal element computed as $D_{ii}=\sum\nolimits_{j = 1}^n {{S_{ij}}} $.

Next, we present an alternating algorithm to optimize the model in Eq.~(\ref{eq:zsh2}).

\subsubsection{Update P}
Fixing all variables except for $P$, we get the quadratic problem as:
\begin{equation}
\min\limits_P \alpha\|{P^T\phi(X) \!- \!B}\|_F^2\!+\!\beta \left\| {P} \right\|_F^2\!+\!\gamma Tr(P^T\phi(X)L\phi(X)^TP).
\label{eq:P}
\end{equation}
By setting its derivative with respect to $P$ to 0, we have the following solution
\begin{equation}
P = \left(\phi(X)\phi(X)^T+\frac{\beta}{\alpha}I+\frac{\gamma}{\alpha}\phi(X)L\phi(X)^T\right)^{-1}\phi(X)B^T.
\label{eq:P}
\end{equation}

\subsubsection{Update B}
In this step, we fix all other variables and learn binary codes $B$ with discrete constraint. The objective function can be reduced to
\begin{equation}
\begin{split}
\mathop {\min }\limits_{B}~&{\left\| {R^TY - {W^T}B} \right\|_{F}^{2}} + \alpha \left\| {P^T\phi(X) - B} \right\|_F^2\\
\textrm{s.t.}~&B\in \{-1,1\}^{l\times n}.
\end{split}
\end{equation}
The above equation can be further written as
\begin{equation}
\begin{split}
\mathop {\min }\limits_{B}~&{\left\| W^TB \right\|_{F}^{2}} -2Tr(B^TH)\\
\textrm{s.t.}~&B\in \{-1,1\}^{l\times n},
\end{split}
\end{equation}
where $H = WR^TY + \alpha P^T\phi(X)$.

Inspired by~\cite{wu2008coordinate}, we apply the \textit{discrete coordinate descent (DCC)} algorithm to solve the above sub-problem. Denote $B$ as $B = \left [q_1^T,q_2^T,\cdots,q_l^T \right ]$, $H = \left [ h_1^T,h_2^T,\cdots,h_l^T \right ]$ and $W = \left [ u_1^T,u_2^T,\cdots,u_l^T \right ]$, where $q_i^T$, $h_i^T$ and $u_i^T$ are the $i$-th row of $B$, $H$ and $W$, respectively. Furthermore, for convenience, we denote
\begin{equation}\left\{ \begin{array}{l}
{{B}_{\neg i}} = \left[ {{q_1}^T,...,{q_{i - 1}}^T,{q_{i + 1}}^T,...,{q_l}^T} \right],\\
\\
{H_{\neg i}} = \left[ {{h_1}^T,...,{h_{i - 1}}^T,{h_{i + 1}}^T,...,{h_l}^T} \right],\\
\\
{{W}_{\neg i}} = \left[ {{u_1}^T,...,{u_{i - 1}}^T,{u_{i + 1}}^T,...,{u_l}^T} \right].
\end{array} \right.
\end{equation}
Then we can have
\begin{equation}
\begin{split}
\| W^TB \|^2 &= Tr(B^TWW^TB)\\
&=const + \| q_i u_i^T\|_F^2+2u_i^TW_{\neg i}^TB_{\neg i}q_i\\
&=const + 2u_i^TW_{\neg i}^TB_{\neg i}q_i.
\end{split}
\label{eq:WB}
\end{equation}
Here, $\|q_i u_i^T\|^2 = Tr(u_i q_i^Tq_iw_i^T) = const$. Following the same rule, we also have the following conclusion
\begin{equation}
Tr(B^TH) = const + h_i^Tq_i.
\label{eq:BQ}
\end{equation}
The sub-problem can be transformed to
\begin{equation}
\begin{split}
\min\limits_{q_i}~&(u_i^TW_{\neg i}^TB_{\neg i}-h_i)q_i\\
\textrm{s.t.}~&h_i\in\{-1, 1\}^{n\times 1},
\end{split}
\end{equation}
The optimal solution of above equation is
\begin{equation}
\begin{split}
q_i = \textrm{sgn}(h_i-B_{\neg i}^TW_{\neg i}u_i).
\end{split}
\label{eq:B}
\end{equation}
where $\textrm{sgn}(\cdot)$ is the sign function. We can see that each bit of the desired binary code $B$ can be learned based on other $l-1$ bits. Thus, we can use cyclic coordinate descent approach to generate the optimal codes until the entire procedure converges.

\subsubsection{Update R}
With $B,W,P$ fixed, we then have
\begin{equation}
\begin{split}
\min\limits_{R}~&\left\| {R^TY - {W^T}B} \right\|_{F}^{2}\\
\textrm{s.t.}~&R^TR = I,
\end{split}
\label{eq:R}
\end{equation}
which can efficiently solved by the algorithm proposed in~\cite{wen2013feasible}.

\subsubsection{Update W}
By fixing $P,B,R$, we arrive at a classic ridge regression problem:
\begin{equation}
\begin{split}
\mathop {\min }\limits_{W}~&{\left\| {R^TY - {W^T}B} \right\|_{F}^{2}} + \lambda {\left\| W \right\|_{F}^{2}}.
\end{split}
\end{equation}
The above equation has an closed-form solution
\begin{equation}
\begin{split}
W = (BB^T+\lambda I_l)^{-1}BY^TR.
\end{split}
\label{eq:W}
\end{equation}
where $I_l$ is a diagonal matrix of size $l\times l$.

By iteratively updating $P,W,B,R$ until convergence, we can arrive at an optima. The overall algorithm is summarised in Algorithm \ref{alg:1}.

\begin{algorithm}[!h]
\begin{algorithmic}[1]
\renewcommand{\algorithmicrequire}{\textbf{Input:}}
\renewcommand{\algorithmicensure}{\textbf{Output:}}
\REQUIRE Training data $X$ and the embedded label matrix $Y$;
\ENSURE Binary code $B$, alignment matrix $R$, hash function $P$ and mapping matrix $W$;
\STATE Randomly initialize $B,P$ and $W$;
\STATE Randomly initialize $R$ to to be orthogonal;
\STATE Map $X$ to $\phi(X)$ using $m$ anchors randomly selected from $X$;
\STATE Construct Laplacian matrix $L$;
\REPEAT{
\STATE Update $P$ according to Eq.~(\ref{eq:P});
\STATE Update $B$ iteratively by using the solution of (\ref{eq:B});
\STATE Update $R$ according to Eq.~(\ref{eq:R});
\STATE Update $W$ according to Eq.~(\ref{eq:W});
}
\UNTIL{there is no change to $P,B,R,W$}
\RETURN{$B,P,W,R$;}
\end{algorithmic}
\caption{Algorithm for optimizing Zero-Shot Hashing}
\label{alg:1}
\end{algorithm}

\subsection{Algorithm Analysis}
In this section, we analyze the convergence and time complexity of our algorithm.

\subsubsection{Convergence Study}
\begin{figure}[t]
  \centering
  \subfloat{
  \includegraphics[width=0.8\linewidth]{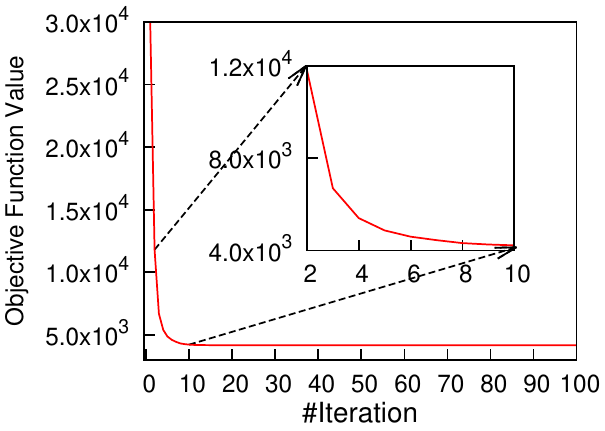}
  }
\caption{Convergence study on ImageNet.}
\label{fig:convergency}
\end{figure}

As shown in Algorithm~\ref{alg:1}, in each iteration, the updates of all variables make the objective function value decreased. We also conducted empirical study on the convergence property using ImageNet~\cite{deng2009imagenet}. Specifically, we trained our zero-shot hashing model with $30,000$ seen images randomly sampled from the ImageNet dataset, with label embedding as supervision. We selected $1,000$ anchors and set the code length to be 64 bits. As Figure \ref{fig:convergency} shows, our algorithm starts with cost function value roughly at $30,000$, but descends dramatically within only 10 iterations, and reaches a stable local minima at the 20-th iteration. This phenomenon clearly indicates the efficiency of our algorithm.

\subsubsection{Computational Complexity}
In each iteration (line 6-9), the time cost is analyzed as follows. The computation of $P$ in Eq.~(\ref{eq:P}) is $O(m^2n+nml+m^3)$. The DCC algorithm for updating $B$ costs $O(cl^2+l^2n)$. As to the optimization of the sub-problem in Eq.~(\ref{eq:R}), the time cost is $O(c^3)$. Finally, the computational cost of updating $W$ is $O(l^2n+lnc+lc^2+l^3)$. Given that $m\ll n$, $l\ll n$, $c\ll n$ and our algorithm converges within a few iterations (less than 10), the overall time cost of our algorithm is $O(n)$. It is worth noting that the dominant operation of our algorithm is matrix multiplication, which can be greatly speeded up by using parallel and/or distributed algorithms.

\section{EXPERIMENT}
\subsection{Experimental Settings}
In our experiments, we employ three real-life image datasets, including CIFAR-10\footnote{https://www.cs.toronto.edu/~kriz/cifar.html}, ImageNet\footnote{http://image-net.org/} and
MIRFlickr\footnote{http://press.liacs.nl/mirflickr/}.

\indent\textbf{CIFAR-10} consists of $60,000$ images which are manually labelled with 10 classes including \textit{airplane, automobile, bird, cat, deer, dog, frog, horse, ship, truck}, with $6,000$ samples in each class. The classes are completely mutually exclusive, \textit{i.e.}, no overlap between classes (\emph{e.g.}, automobiles and trucks).

\textbf{ImageNet} is an image dataset organized according to the WordNet~\cite{miller1995wordnet} hierarchy. The subset of ImageNet for the Large Scale Visual Recognition Challenge 2012 (ILSVRC2012) is used for our experiments, consisting of over 1.2 million Web images, manually labeled with $1,000$ object categories.

\textbf{MIRFlickr} consists of $25,000$ images collected from the social photography site Flickr through its public API. Firstly introduced in 2008, this dataset is wildly used in multimedia research. MIRFlickr is a multi-label dataset with every image associating with 24 popular tags such as \textit{sky}, \textit{river}, \textit{etc}.

For all image data, we adopted the winning model for the 1000-class ImageNet Large Scale Visual Recognition Challenge 2012 ~\cite{krizhevsky2012imagenet} to extract the fully connected layer fc-7 as visual feature.

Various metrics are employed for performance of different evaluation tasks. For image retrieval, we used the two traditional metrics \emph{i.e.}, Precision and Mean Average Precision (MAP). MAP focuses on the ranking of retrieval results and we reported the results over the top $5,000$ retrieved samples. Precision mainly concentrates on the retrieval accuracy and we reported the results with Hamming radius $r\leq2$.

We compared our proposed ZSH with four state-of-the-art supervised hashing approaches, including COSDISH~\cite{kang2016column}, SDH~\cite{shen2015supervised}, KSH~\cite{liu2012supervised} and LFH~\cite{zhang2014supervised}. For all anchor-based algorithms, we randomly sampled $1,000$ anchors from the training dataset. Furthermore, we compared to one of the most representative unsupervised hashing method, \emph{i.e.}, Inductive Hashing on Manifolds (IMH)~\cite{shen2013inductive}.

For all comparing approaches, we followed their suggested parameter settings. For ZSH, we empirically set $\alpha$ to $10^{-5}$ and $\gamma$ to $10^{-6}$. For regularization parameters $\lambda$ and $\beta$, we set them to $10^{-2}$ and $10^{-4}$, respectively. The number of iterations is set to 10. We define the similarity matrix $S$ to be computed by
\begin{equation}\nonumber
{S_{ij}} = \left\{ {\begin{array}{*{20}{c}}
{\exp (\frac{{{{\left\| {{x_i} - {x_j}} \right\|}_2}}}{{2{\sigma ^2}}}),}&{{\rm{if }}~{x_i} \in {\mathcal{N}_k}({x_j})~{\rm{ or }}~{x_i} \in {\mathcal{N}_k}({x_j})}\\
\\{0,}&{{\rm{otherwise,}}}
\end{array}} \right.
\end{equation}
where $\mathcal{N}_k(\cdot)$ is the function of searching $k$ nearest neighbors. In our experiment, we set $\sigma = 1$.

\subsection{Results on CIFAR-10}
\subsubsection{Overall Comparison of Zero-shot Image Retrieval}
\begin{figure*}[!ht]
  \centering
  \includegraphics[width=1\linewidth]{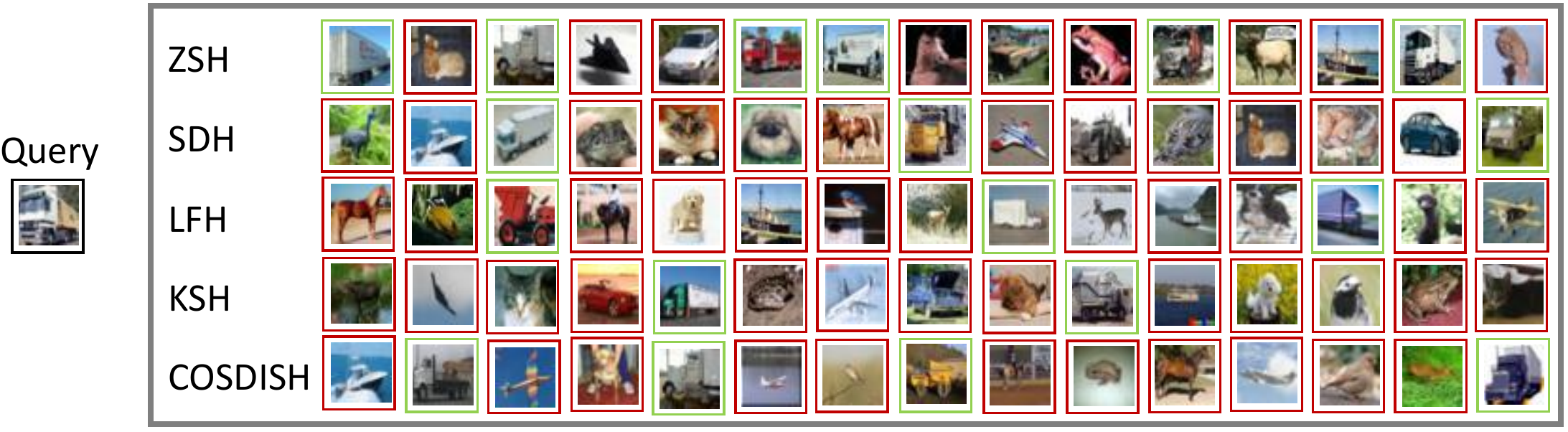}\\
  \caption{A demonstration of the zero-shot image retrieval exemplars using different comparing hashing algorithms on CIFAR-10, where top $15$ retrieval images are reported. Pictures with green bounding box indicate the correct results while those with red outlines indicate failure results. As can be seen, our proposed ZSH method returns the largest number of correct retrieved results with query from unseen category, followed by COSDISH which returns four correct samples.}
\label{fig:retrieval}
\end{figure*}

To evaluate the efficacy of retrieving images in unseen categories, we split CIFAR-10 into a ``seen'' training set and an ``unseen'' testing set. In particular, we select \textit{truck} as unseen testing category and leave the rest 9 categories as seen training set. For all comparing algorithms, we randomly sample $10,000$ images for learning hash functions. For testing purpose, we randomly select $1,000$ images from the unseen category as query images, and the remaining $5,000$ test images together with the $54,000$ images of seen categories are combined to form the retrieval database.

\begin{figure}[htb!]
  \centering
  \includegraphics[width=1\linewidth]{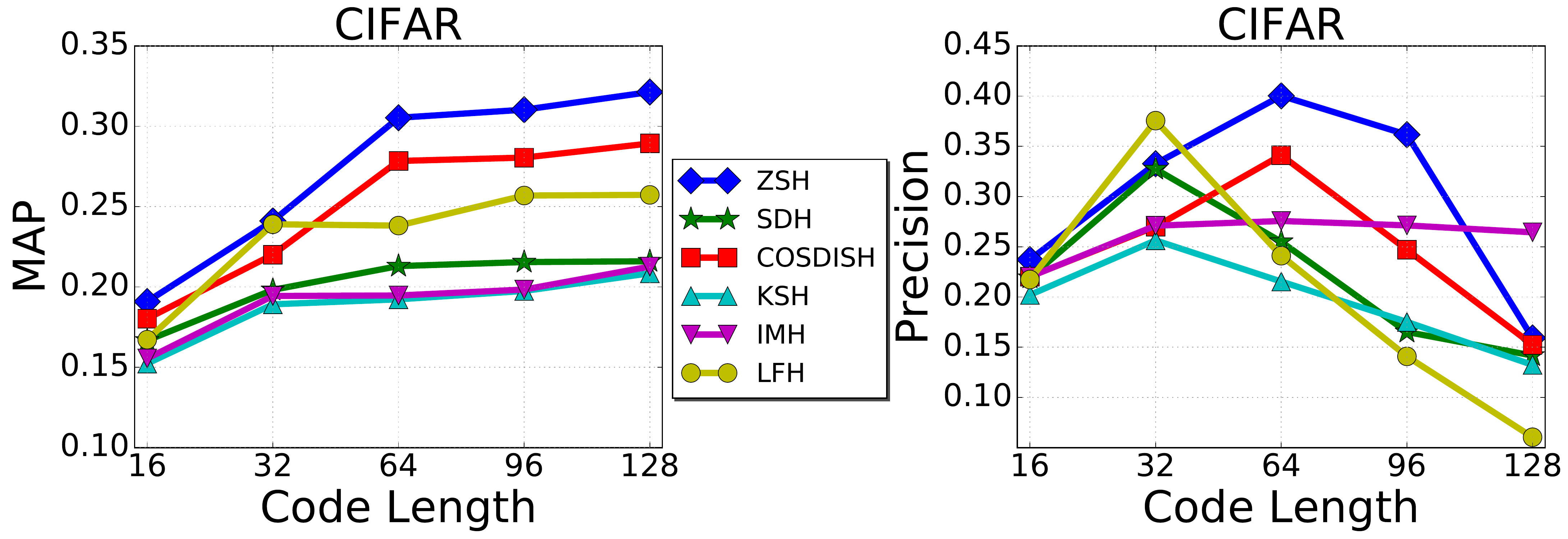}
   \caption{Performance (MAP and Precision) of different comparing methods on zero-shot image retrieval over CIFAR-10 dataset.}
  \label{fig:cifar_zeroshot}
\end{figure}

The performance of all comparing approaches \emph{w.r.t.} different codes lengths (\emph{i.e.}, $\{16, 32, 64, 96, 128\}$) is illustrated in Figure~\ref{fig:cifar_zeroshot}. As we can see, the proposed ZSH outperforms all the other hashing algorithms in terms of MAP at all code lengths. As to Precision, ZSH still shows superior image retrieval performance in most cases. The underlying principle is that our method not only utilizes inherent semantic relationship among labels to transfer supervisory knowledge, but also preserves discrete and structural properties of data in the learning of hash codes and hash functions. An interesting observation is the performance of IMH, which is an unsupervised method, gains competitive even better retrieval results in terms of Precision as compared to some supervised methods such as KSH, SDH. While unsupervised methods encode images solely with the distributional properties in the feature space, the supervised ones may be misled by independent semantic labels in the learning processing.

Besides, MAP increases rapidly for all methods when code length varies from $16$ to $64$, and then reaches a slow-growth stage from 64 bits to 128 bits. When code length is short, more codes are required to guarantee the descriptive and discriminative power. However, after encoding space is large enough (\textit{e.g.} 64 bits), the expression ability saturated, providing more bits cannot significantly improve the performance. As to Precision, hashing performance significantly deteriorates as code length is larger than $64$. Recall that our searching radius is empirically set to 2, forming a hyper-ball of radius 2 in Hamming space. When the code length increases from $16$ to $64$, significant improvement in retrieval ability counteracts the searching difficulty. However, as Hamming space becomes larger, searching difficulty grows linearly, thereby degrading the Precision performance. Therefore, as a trade-off between efficiency and effectiveness, an eclectic code length should be chosen.

\subsubsection{Effect of Different Unseen Category}
\begin{figure}[htb!]
  \centering
  \includegraphics[width=1\linewidth]{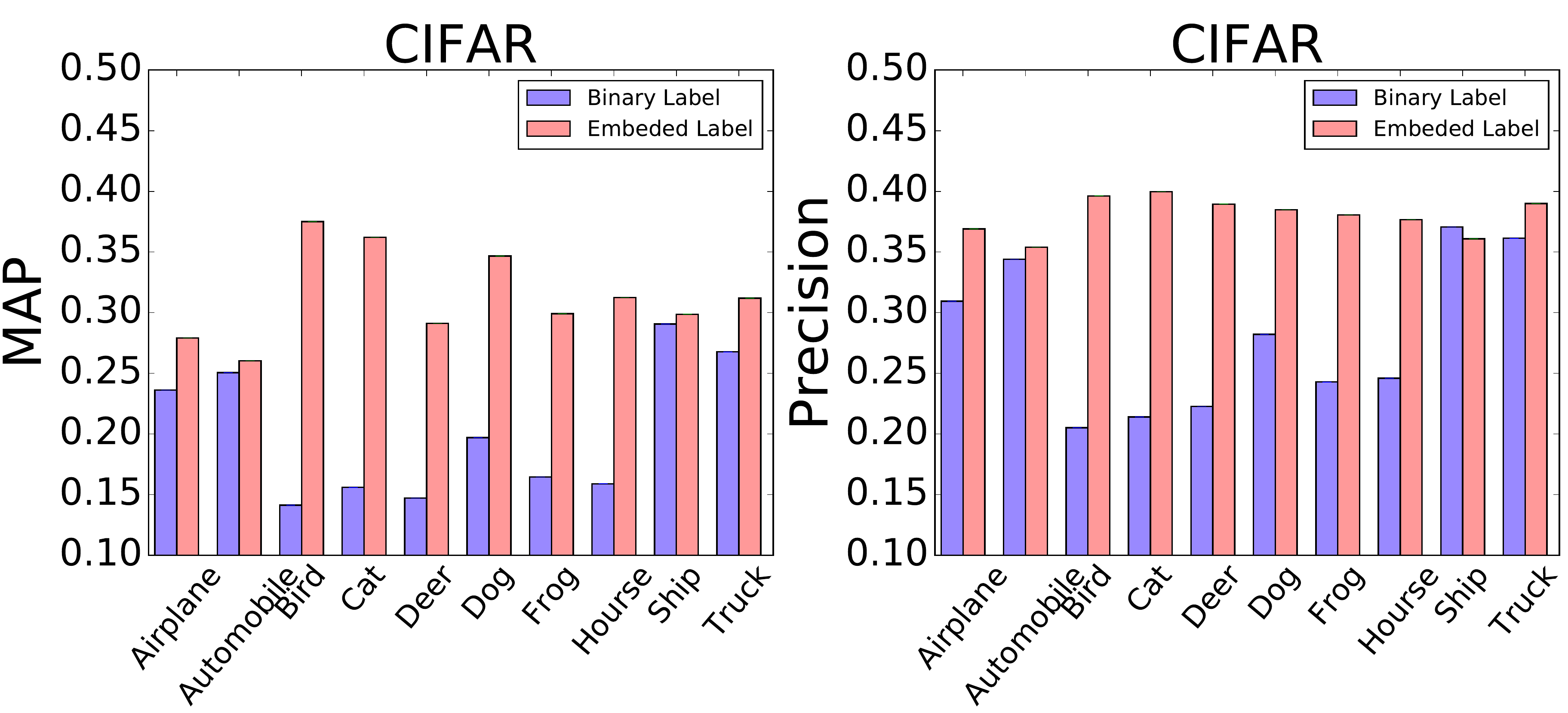}
  \caption{Performance (Precision and MAP) of zero-shot image retrieval for each individual unseen category on CIFAR-10 dataset.}
   \label{fig:cifar_fig9}
\end{figure}

In this experiment, we aim to evaluate the performance of zero-shot image retrieval on different unseen categories. The experimental settings are the same as that in the previous subsection. Figure~\ref{fig:cifar_fig9} illustrates the MAP and Precision performance of ZSH using each individual label as unseen testing data.

We can observe that zero-shot image retrieval performance varies from one class to another, reaching peak at \textit{bird} and bottom at \textit{automobile}. Intuitively, if an unseen class is semantically closer to other seen categories, more relevant supervisory knowledge can be transferred from word embedding space for boosting the retrieval performance. To dig deeper about the reason behind the fluctuation of performance on different unseen objects, we compute the average cosine similarity between each unseen category and other seen categories, and list the corresponding MAP in Table \ref{Tab:sim}.

\begin{table}[!h]
\centering
\begin{tabular}{lcc}
\hline
\textbf{Category} & \textbf{Average Cosine Similarity} & \textbf{MAP}\\
\hline
airplane & 0.2191 & 0.2791
\\ automobile & 0.1567 & 0.2603
\\ bird & 0.3565 & 0.3751
\\ cat & 0.3661 & 0.3621
\\ deer & 0.2981 & 0.2912
\\ dog & 0.3826 & 0.3467
\\ frog & 0.3485 & 0.2991
\\ horse & 0.3015 & 0.3125
\\ ship & 0.1663 & 0.2987
\\ truck & 0.3358 & 0.3120
\\
\hline
\end{tabular}
\caption{Average cosine similarity of each category and all other categories, together the corresponding MAP performance.}
\label{Tab:sim}
\end{table}

We observe that the MAP performance is positively related to the average cosine similarity. For instance, those of larger cosine similarity (\textit{e.g.}, \textit{dog, cat}) performs relatively well, while those of smaller similarity (\textit{e.g.}, \textit{airplane, automobile}) gain relatively poor performance. This observation implies that in order to achieve satisfactory retrieval results, unseen classes should have sufficient correlation with seen ones.

As shown in Figure~\ref{fig:cifar_fig9}, we also compare the effects of embedded labels and binary labels. The performance of embedded labels is obviously better than that of binary labels. The underlying reason is that the embedding space can help to capture the relationship between seen and unseen categories for transferring supervisory knowledge. In contrast, binary labels neglect semantic correlations, thereby leading to irregular fluctuations of retrieval performance.

\subsubsection{Effect of Seen Category Ratio}
In this experiment, we evaluate the performance of our proposed ZSH \emph{w.r.t.} different numbers of seen categories. Specifically, we vary the ratio of seen categories in the training set from $0.1$ to $0.9$. For each ratio, we randomly sample $10,000$ images from the seen categories for training. Further, we randomly select $1,000$ images from the unseen set as queries to search in the remaining $59,000$ images. Note that when the ratio of seen categories decreases to $0.1$, we use all $6,000$ datums of that class as training set.

\begin{figure}[!h]
  \centering
  \includegraphics[width=1\linewidth]{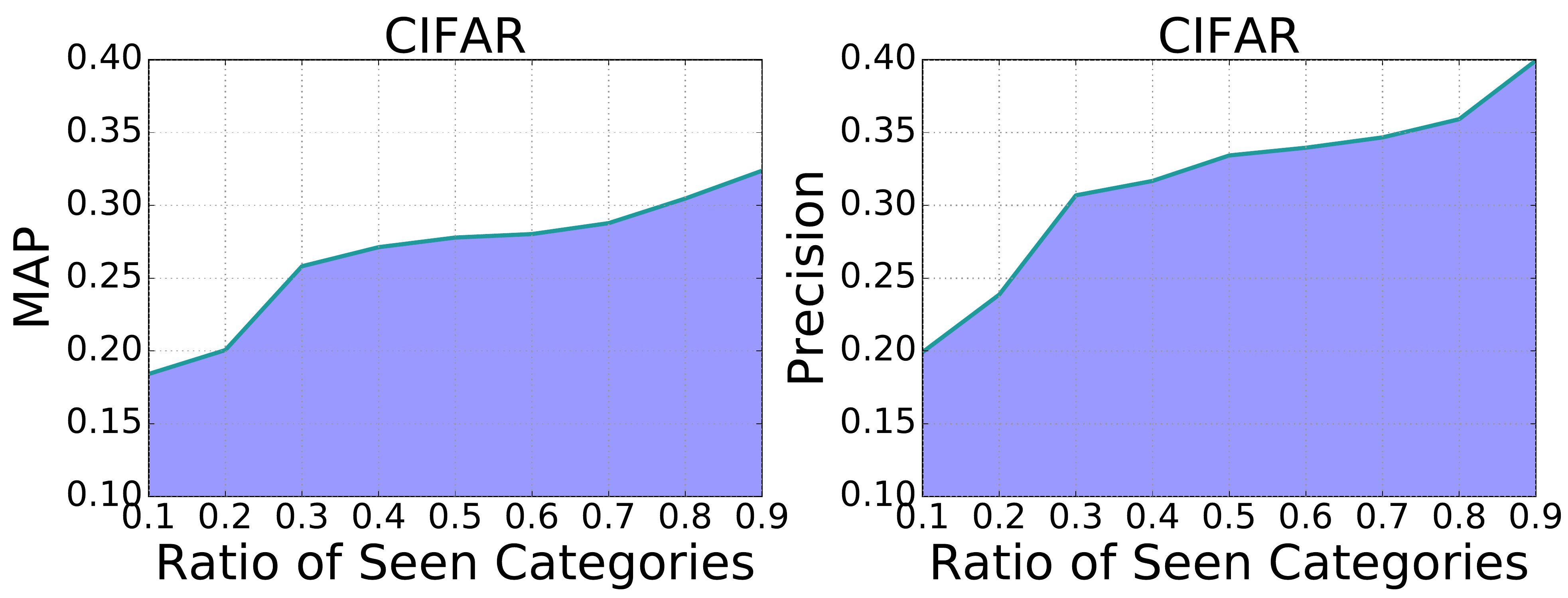}
  \caption{Effects of different ratios of seen categories on CIFAR-10 dataset.}
  \label{fig:unseen_ratio}
\end{figure}
We report the experimental results in Figure \ref{fig:unseen_ratio}, from which we have the following observations: (1) The performance of both MAP and Precision boosts as the ratio of the seen categories grows; (2) As the ratio increases from $0.1$ to $0.3$, we see a dramatic leap of the retrieval performance, followed by a relatively slight performance improvement from $0.3$ to $0.9$. We analyze that by observing more ``seen'' categories, we have higher possibility to find relevant supervision for the unseen class, which guides to learn better intermediate hash codes, thereby simultaneously improving the quality of hash functions.

\subsubsection{Effect of Training Size}
\begin{figure}
  \centering
  \includegraphics[width=0.8\linewidth]{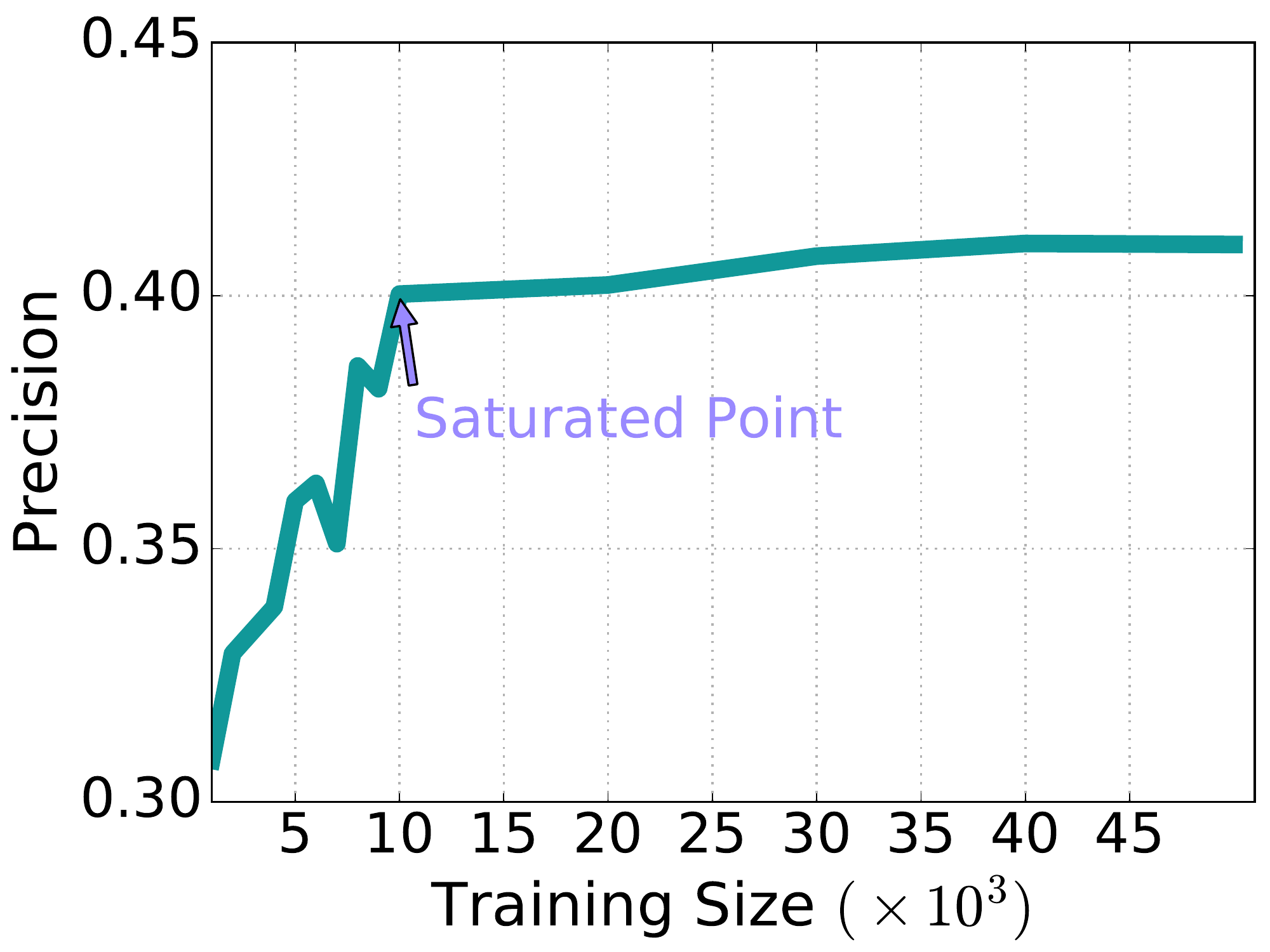}
\caption{Effects of training size on Precision performance over CIFAR-10 dataset.}
\label{fig:trainingsize}
\end{figure}

This part of experiment mainly focuses on evaluating the effect of training size on the searching quality of ZSH. We select \textit{truck} as the unseen object and varies the size of training data in the range of $\{1000,2000,\ldots,10000,20000,\ldots,50000\}$. The results are demonstrated in igure~\ref{fig:trainingsize}. As we can see, when the size increases from $1,000$ to $10,000$, we observe a rapid rise of the Precision performance. Nonetheless, when fed with more training data, ZSH does not gain noticeable performance boost. For the balance of training efficiency and effectiveness, in the rest experiments, we consistently set the training size to $10,000$.

\subsection{Results on ImageNet}
\subsubsection{Overall Comparison of Zero-shot Image Retrieval}
In this part, we evaluate our proposed ZSH on zero-shot image retrieval as compared to other state-of-the-art methods using the Large Scale Visual Recognition Challenge 2012 (ILSVRC2012) dataset. Recall that the ILSVRC2012 dataset contains more than $1.2$ million images tagged with $1,000$ synsets without any overlap. For evaluation purpose, we randomly choose $100$ categories which have corresponding word embedding learned from Wikipedia text corpus, which gives us a set of roughly $130,000$ images. We split the data into a training set ($90$ seen categories) and a testing set ($10$ unseen categories). For all comparing algorithms, we randomly select $10,000$ images of seen categories for training. As to image queries, we randomly sample $1,000$ images from the unseen categories. We use the learned hash function to encode all the remaining images to form the retrieval database.

\begin{figure}[t]
  \centering
  \includegraphics[width=1\linewidth]{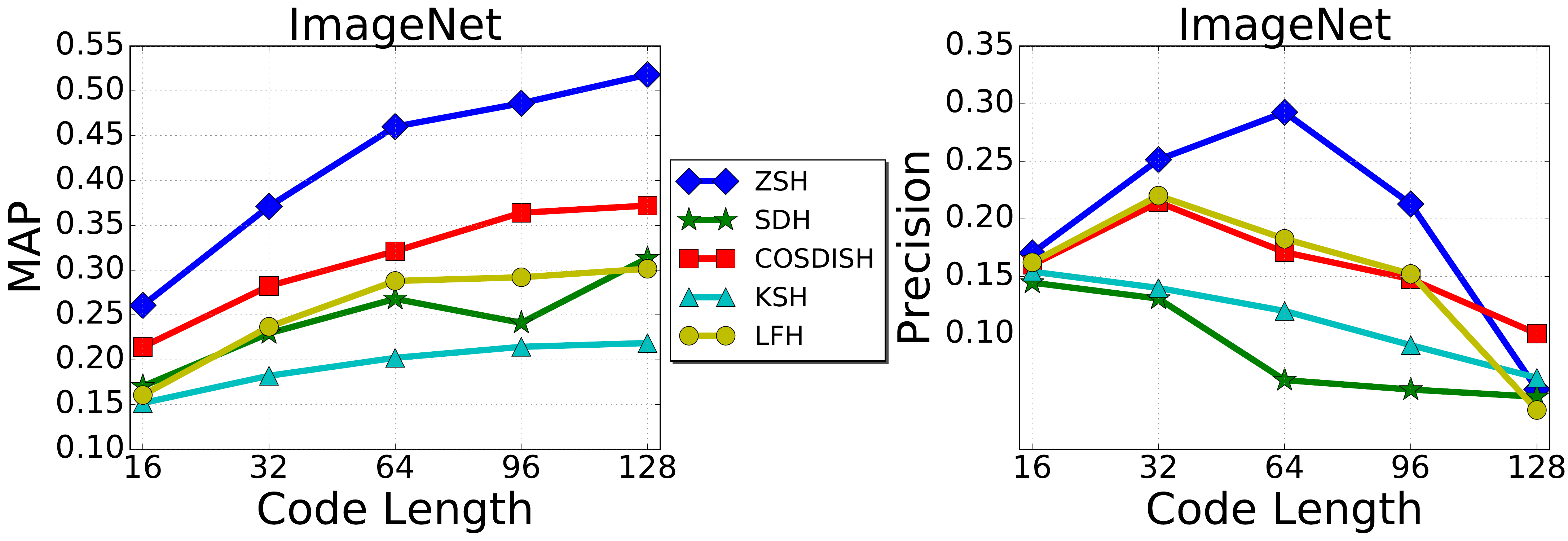}
  \caption{Performance (MAP and Precision) of different comparing methods on zero-shot image retrieval over ImageNet dataset.}
  \label{fig:Imagenet}
\end{figure}

The performance of our proposed ZSH and other four state-of-the-art supervised hashing methods with different code lengths are reported in Figure \ref{fig:Imagenet}. As we can see, ZSH consistently outperforms all other competitors in most cases. As code length varies from $16$ to $128$, we can observe the similar variation tendency of performance on ImageNet to that on CIFAR-10. This phenomenon again implies that we should choose a trade-off code length to guarantee the retrieval performance.

\subsubsection{Image Retrieval in Related Categories}

In zero-shot image retrieval scenario, we expect that even though we fail to retrieve relevant images of the same category, we can still obtain semantically related images. For instance, if the query image describes a \textit{cat}, we may prefer to retrieve images of \textit{dog} rather than images of \textit{car}. Our proposed ZSH utilizes semantic embedding to set up connections between semantically similar labels in the embedded space. In this way, the supervision knowledge of seen categories can be transferred into hash functions, which can effectively encode images of unseen categories.
\begin{figure}[!h]
  \centering
  \includegraphics[width=1\linewidth]{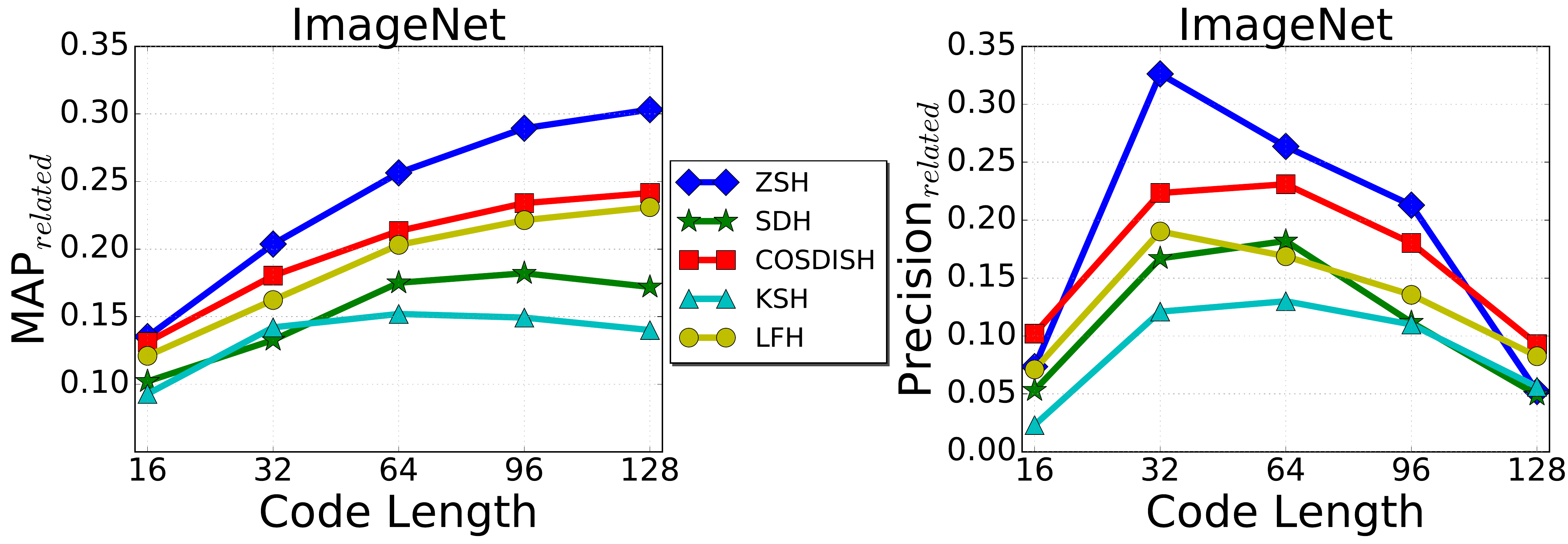}
  \caption{Comparison of ZSH and other hashing approach on the capability of retrieving semantically similar images from related categories.}
  \label{fig:Imagenet_hier}
\end{figure}

Since we need to search more related categories, all remaining images of both seen and unseen categories are used to form retrieval database. All the other settings are the same as that in Section 4.3.1. In order to evaluate the performance of retrieving related categories, we use two modified metrics, named MAP$_{{related}}$ and Precision$_{{related}}$, which are defined as
\begin{equation}
\begin{split}
\textrm{MAP}_{related} = \sum_{i=1}^{K}\frac{\textrm{AP} _{related}@i}{K},
\end{split}
\end{equation}
\begin{equation}
\begin{split}
\textrm{Precision}_{related} = \frac{n_{related}} {n_{retrieved}} ,
\end{split}
\end{equation}

where MAP$_{{related}}$ is calculated based on the top $K$ retrieved results, $AP_{related}@i$ is the average precision based on the related results, calculated by
\begin{equation}
\begin{split}
\textrm{AP}_{related}@i = \frac{n_{related}^{(i)}} {i} ,
\end{split}
\end{equation}
 where $n_{related}^{(i)}$ is the number of related images in top $i$ retrieved results. $n_{related}$ and $n_{retrieved}$ are the related retrieval under Hamming radius 2 and total examples retrieved under Hamming radius 2, respectively. Using WordNet~\cite{miller1995wordnet}, which is a lexical database for the English language, we define query $A$ and retrieved object $B$ are related if: 1) $A$ and $B$ are not of the same category. 2) $A$ can reach $B$ on WordNet within 5 hops.

In practice, we set $K=5,000$ and $R=2$. Figure \ref{fig:Imagenet_hier} shows the experimental results. We can see that in terms of MAP$_{related}@K$, our method always outperforms other methods at every code length. When we look at Precision$_{related}$, our proposed ZSH achieves $0.3262$, $0.2636$, $0.2129$ at 32 bits, 64 bits and 96 bits, which significantly outperforms the second best method. This observation indicates that ZSH is capable of detecting the semantically similar images from the most related categories.

\subsection{Results on MIRFlickr}

\begin{figure}[!h]
  \centering
  \includegraphics[width=1\linewidth]{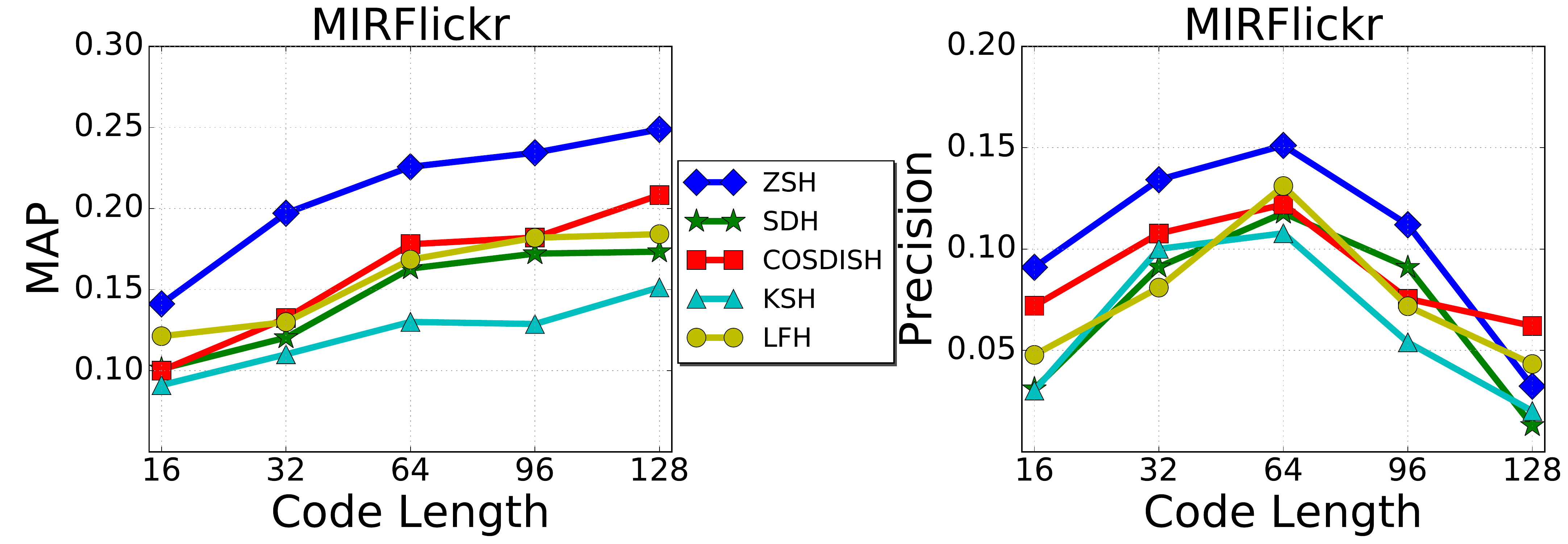}
  \caption{Performance (MAP and Precision) of different comparing methods on zero-shot image retrieval over MIRFlickr dataset.}
   \label{fig:flickr}
\end{figure}
In real-world pictures, especially in user-generated photos, there often exists multiple tags belong to one picture. To examine the practical efficacy of our proposed ZSH, in this part, we conduct an extra experiment on a real-life multi-label dataset, \emph{i.e.}, MIRFlickr, which contains $25,000$ images downloaded from the social photography site Flickr. Each image is associated with $24$ tags. Since in multi-label image dataset, different categories share overlapping images, which makes it difficult to divide the dataset into training set and testing set. Hence, we employ ImageNet as an auxiliary dataset to train our hash functions and evaluate the zero-shot image retrieval performance on MIRFlickr. Specifically, from the ILSVRC2012 dataset we select $100$ categories which does not overlap with the $24$ tags in MIRFlickr. For fair comparison, all hashing approaches use $10,000$ randomly sampled images for training. After the hash function is learned, we directly apply them to transform the MIRFlickr images into binary codes. We then sample $1,000$ datums as query images and search in the remaining $24,000$ images. We regard the retrieval images sharing at least two tags with the query as the true neighbors, and compute MAP on the top $5,000$ retrieved results and Precision under Hamming distance 2. Figure \ref{fig:flickr} illustrates our results of our ZSH and other comparing algorithms on MIRFlickr. In the left sub-figure, we can see that with different code lengths, our ZSH can consistently achieve the best MAP performance among all the comparing algorithms. As the code length increases, the MAP performance of each algorithm keeps increasing, reaching $0.2488$ at $128$ bits, which outperforms the second best hashing method COSDISH by $19\%$ at the same length. In terms of Precision, ZSH exceeds all other methods in most cases. Similar to that of CIFAR-10 and ImageNet, we can see a variation pattern with an increasing trend from $16$ to $64$ and a performance drop from $64$ to $128$. The promising performance on MIRFlickr demonstrates the potential of ZSH in indexing and searching real-life image data.

\section{CONCLUSION}
With the explosion of newly-emerging concepts and multimedia data on the Web, it is impossible to supply existing supervised hashing methods with sufficient labeled data in time. In this paper, we studied the problem of how to map images of unseen categories using hash functions learned from limited seen classes. We proposed a novel hashing scheme, termed \emph{zero-shot hashing} (ZSH), which is capable of transmit supervised knowledge from seen categories to unseen categories. Independent $0/1$-form labels were projected into an off-the-shelf embedding space with abundant semantics, where label semantic correlations can be fully characterized and quantified. Considering the issues of domain difference and semantic shift, we further narrowed down the gap between binary codes and high-level semantics by a semantic alignment operation. Specifically, we intentionally rotated the embedding space to adjust the supervised knowledge more suitable for learning high-quality hash codes. Besides, we also preserved local structural property and discrete nature of hash codes in the ZSH model. An effective algorithm was designed to optimize the model in an iterative manner and the empirical study showed the convergency and efficiency. We evaluated our proposed ZSH hashing approach on three real-world image datasets, including CIFAR-10, ImageNet and MIRFlickr. The experimental results demonstrated the superiority of ZSH as compared to several state-of-the-art hashing approaches on the zero-shot image retrieval task.

In the future, we plan to enhance the exploration of label semantic correlations by integrating knowledge from multiple sources, including textual corpus and visual clues. We expect this will compensate the incomplete representation of each individual modality, thereby solving the problem of domain difference and semantic shift fundamentally.
\bibliographystyle{abbrv}

\bibliography{sigproc}
\end{document}